\documentclass[runningheads]{llncs}
\usepackage{graphicx}
\usepackage{comment}
\usepackage{amsmath,amssymb} 
\usepackage{color}

\usepackage{xspace}
\usepackage{array,multirow}
\usepackage{ mathrsfs }
\usepackage{amssymb}
\usepackage[dvipsnames]{xcolor}
\usepackage{amsmath}
\usepackage{dsfont}
\usepackage{pifont}

\usepackage{color}
\usepackage{xspace}
\usepackage{url}
\usepackage[subtle]{savetrees} 
\usepackage{silence} 
\usepackage[font={footnotesize},labelfont=bf]{caption,subfig} 

\let\llncssubparagraph\subparagraph
\let\subparagraph\paragraph
\usepackage[compact]{titlesec}  
\titlespacing*{\section}
{0pt}{.5ex plus .5ex minus .1ex}{.5ex plus .1ex}
\titlespacing*{\subsection}
{0pt}{0ex plus .5ex minus .1ex}{0ex plus .1ex}
\let\subparagraph\llncssubparagraph

\usepackage{enumitem}

\usepackage{wrapfig}

\renewenvironment{equation}{\vspace{-1em}\begin{oldequation}}{\end{oldequation}\vspace{-1em}}

\usepackage{booktabs} 

\usepackage{array}
\newcommand{\PreserveBackslash}[1]{\let\temp=\\#1\let\\=\temp}
\newcolumntype{C}[1]{>{\PreserveBackslash\centering}p{#1}}
\newcolumntype{R}[1]{>{\PreserveBackslash\raggedleft}p{#1}}
\newcolumntype{L}[1]{>{\PreserveBackslash\raggedright}p{#1}}
\newcolumntype{M}[1]{>{\centering\arraybackslash}m{#1}}

\newcommand{\sys}[0]{Adversarially-Trained Autoencoder Augmentation\xspace} 
\newcommand{\sysshort}[0]{AAA\xspace} 
\newcommand{\lossx}[0]{$\mathbf{L}_{xent}$\xspace}
\newcommand{\lossm}[0]{$\mathbf{L}_{mse}$\xspace}
\newcommand{\lossxm}[0]{$\mathbf{L}_{xent} + \mathbf{L}_{mse}$\xspace}

\newcommand{\sysx}[0]{$\sysshort_{\mathbf{L}_{xent}}$\xspace}
\newcommand{\sysm}[0]{$\sysshort_{\mathbf{L}_{mse}}$\xspace}
\newcommand{\sysxm}[0]{$\sysshort_{\mathbf{L}_{xent} + \mathbf{L}_{mse}}$\xspace}

\newif\ifsubmit

\ifsubmit
    \newcommand{\amir}[1]{}
        \newcommand{\kevin}[1]{}
    \newcommand{\pratik}[1]{}
    
    \newcommand{\todo}[1]{}
\else
    \newcommand{\amir}[1]{\textcolor{blue}{Amir: #1}}
    \newcommand{\kevin}[1]{\textcolor{green}{Kevin: #1}}
    \newcommand{\pratik}[1]{\textcolor{cyan}{Pratik: #1}}
    
    \newcommand{\todo}[1]{\textcolor{red}{TODO: #1}}
\fi

\newcommand{\paratitle}[1]{\noindent\textbf{\textit{#1.}}\xspace}

\newcommand{\ie}[0]{\emph{i.e.,}\xspace}
\newcommand{\etal}[0]{\emph{et al.}\xspace}
\newcommand{\eg}[0]{\emph{e.g.,}\xspace}

\usepackage{cleveref} 

\crefformat{section}{\S#2#1#3} 
\crefformat{subsection}{\S#2#1#3}
\crefformat{subsubsection}{\S#2#1#3}

\begin{document}
\pagestyle{headings}
\mainmatter
\def\ECCVSubNumber{4398}  

\title{Towards Model-Agnostic Adversarial Defenses\\using Adversarially Trained Autoencoders} 

\titlerunning{Towards Model-Agnostic Adversarial Defenses}
%
\author{Pratik Vaishnavi\inst{1} \and
Kevin Eykholt\inst{2} \and
Atul Prakash\inst{3} \and
Amir Rahmati\inst{1}}
\authorrunning{P. Vaishnavi, K. Eykholt, A. Prakash and A. Rahmati}
%
\institute{Stony Brook University, Stony Brook, NY, USA\\
\email{\{pvaishnavi,amir\}@cs.stonybrook.edu} \and
IBM Research, Yorktown Heights, NY, USA\\
\email{kheykholt@ibm.com} \and
University of Michingan, Ann Arbor, MI, USA\\
\email{aprakash@umich.edu} 
}
\maketitle

\begin{abstract}
Adversarial machine learning is a well-studied field of research where an adversary causes predictable errors in a machine learning algorithm through precise manipulation of the input. Numerous techniques have been proposed to harden machine learning algorithms and mitigate the effect of adversarial attacks. Of these techniques, adversarial training, which augments the training data with adversarial samples, has proven to be an effective defense with respect to a certain class of attacks. However, adversarial training is computationally expensive and its improvements are limited to a single model. In this work, we take a first step toward creating a model-agnostic adversarial defense. We propose \sys (\sysshort),  the first model-agnostic adversarial defense  that is robust against certain adaptive adversaries. We show that \sysshort allows us to achieve a partially model-agnostic defense by training a single autoencoder to protect multiple  pre-trained classifiers; achieving adversarial performance on par or better than adversarial training without modifying the classifiers. Furthermore, we demonstrate that \sysshort can be used to create a fully model-agnostic defense for MNIST and Fashion MNIST datasets by improving the adversarial performance of a never before seen pre-trained classifier by at least 45\% with no additional training.  Finally, using a natural image corruption dataset, we show that our approach improves robustness to naturally corrupted images, which has been identified as strongly indicative of true adversarial robustness.

\end{abstract}

\section{Introduction}
\label{sec:intro}
Machine learning algorithms are becoming the preferred tool to empower systems across multiple applications domains ranging from automatically monitoring employee hygiene and safety to influencing control decisions in self-driving cars and trading. With such pervasive use, it is critical to understand and address the vulnerabilities associated with machine learning algorithms so as to mitigate the risks in real systems. Adversarial attacks  are one class of such vulnerabilities in which an adversary can reliably induce predictable errors in machine learning systems. At a high-level, given a classification model and a correctly labelled input, an adversarial attack computes the necessary modifications on the input such that the model incorrectly labels it, while ensuring that the modifications are inconspicuous (\ie imperceptible to a human observer).


To mitigate or prevent the effect of adversarial attacks, multiple defensive techniques have been proposed. Adversarial training is one such technique, which uses a data augmentation strategy to improve a model's performance in adversarial scenarios~\cite{madry2018towards}. During training, adversarial examples are computed on-the-fly and added to the training data. This approach has proven to greatly improve the prediction accuracy on a certain class of adversarial inputs\footnote{We use \textit{class} to denote a group of adversarial examples generated using the same attack algorithm and attack hyper-parameters}. However, like numerous other defense, adversarial training is a \textbf{model-specific defense}. That is to say, the robustness guarantees are limited to the updated model due to invasive, and specific modifications to vulnerable model. In the case of adversarial training, it requires re-tuning the model's parameters to counter adversarial noise. Also, the process of adversarial training introduces significant performance overhead during training as the computationally expensive process of generating adversarial examples must be repeated every training step. Such limitations make adversarial training and other invasive defenses \cite{distillation,ilyas2017robust} undesirable when a model's architecture may change periodically or there is a need for creating multiple robust models.


Concretely, many traditional defenses make a key a assumption: whitebox access to the model. This access is necessary in order to view the internal model state, such as the model's loss gradients and, for some works, deploy the defense by modifying the model's state. Such access is taken for granted as prior work assumes that the user of the machine learning model is the same entity that designs and deploys it. If whitebox access is not possible, then many proposed defenses cannot be deployed. However, when would it be the case that the user of a machine learning algorithm is not the same entity as the designer?

Ji~\etal demonstrate one such case in their paper on model-reuse attacks on machine learning systems \cite{ji2018model}. Due to the growing use of machine learning complex systems, they hypothesized that developers had less time to fully understand and implement their own machine learning architecture and thus, integrated pre-existing, pre-trained architectures in the wild. This model-reuse is further encourage by the existence of open source implementations in multiple frameworks and model zoos. To verify this claim, they examined if active machine learning GitHub repositories in 2016 used one of five popular machine learning models: GoogLeNet, AlexNet, Inception v3, ResNet, and VGG. Of the 16,167 repositories identified, 2200 (13.7\%) of them re-used at least one of the five pre-trained models. Unlike third party software, machine learning models are stateful, which introduces new security and privacy issues involved in re-use. Another situation where whitebox access may not be possible may be in situations where one seeks to update a pre-trained model, but does not have the required visibility due to propriety information or sensitive training data. In these cases, it would be useful to have a \textbf{model-agnostic} defense.

\begin{table*}[tb!]
    \centering
    \resizebox{\textwidth}{!}{%
    \begin{tabular}{@{}M{50mm}M{15mm}M{15mm}M{15mm}M{15mm}M{15mm}M{15mm}@{}}
    \toprule
    \textbf{Properties} & \textbf{Madry \etal~\cite{madry2018towards}} & \textbf{Ilyas \etal~\cite{Ilyas2019AdversarialEA}} & \textbf{Liao \etal~\cite{Liao2018DefenseAA}} & \textbf{Meng \etal~\cite{meng2017magnet}} &  \textbf{Defense GAN~\cite{defensegan}} & \textbf{\sysshort}\\
    \midrule
    Unintrusive & \ding{56} & \ding{56} & \ding{52} & \ding{52} & \ding{52} & \ding{52} \\
    Transferable & \ding{56} & \ding{56} & \ding{52} & \ding{52} & \ding{52} & \ding{52} \\
    Black-box Compatible & \ding{56} & \ding{56} & \ding{52} & \ding{52} & \ding{52}* & \ding{52}* \\
    \cmidrule(lr){1-7}
    Adaptive White-Box Robustness & \ding{52} & \ding{52} & \ding{56} & \ding{56} & \ding{56} & \ding{52} \\
    \bottomrule
    \end{tabular}
    }
\caption{Salient properties of a model-agnostic defense and how they apply to the other types of defenses discussed in this paper, related works and the two variants of \sysshort. \\ \textit{\scriptsize{ *Only compatible for simpler datasets such as MNIST and Fashion-MNIST.}}}
\label{tab:properties}
\vspace{-3em}
\end{table*}

In this paper, we propose a shift toward \emph{model-agnostic defenses}. A fully model-agnostic defense has to meet three key properties:

\begin{enumerate}[nosep]
    \item \textbf{Unintrusive:} It does not modify the internal state of the models it's protecting to achieve adversarial robustness. This includes changing the model weights, number of hidden layers, or size of the input.
    \item \textbf{Transferable:} For a set of pre-trained models, it can be transferred to each model without modification while achieving similar levels of adversarial robustness.
    \item \textbf{Black-box Compatible:} For a set of pre-trained models, it should not require information about the model state beyond its output decision. 
\end{enumerate}

In essence, a fully model-agnostic defense should be able to treat any pre-trained model for a given task as a black-box and improve its adversarial robustness no matter the adversarial threat model. To this end, we introduce \emph{\sys} (\sysshort), a step toward designing a \textbf{model-agnostic} defense against first order adaptive adversarial attacks. To achieve \textit{unintrusiveness}, we re-examine data pre-processing adversarial defenses and revisit the stacked autoencoder (AE) - classifier architecture, in which a denoising AE is used to mitigate the effect of adversarial inputs. Some prior works have also used this approach and successfully created model agnostic defenses with one caveat~\cite{DBLP:journals/corr/GuR14,Liao2018DefenseAA,meng2017magnet}. These prior works trained the architecture to be robust to a static or black-box adversary. The adversarial inputs were generated under the assumption an adversary would not modify their approach to bypass the defense. Later work demonstrated that simple modification to existing attacks would break theses defenses~\cite{carlini2017magnet,athalye2018robustness}. Our approach differs from these previous model-agnostic defenses in that we adversarially train the AE against the aforementioned adaptive adversary using a loss that encourages enhanced classification performance in adversarial scenarios.

To achieve \textit{transferability}, we use ensemble adversarial training, a technique used in prior work for a similar reason \cite{tramer2018ensembletraining}. This modification provides the added benefit that enables parallelizing adversarial training across multiple classifiers to achieve a partially model-agnostic defense. This setup allows \sys to achieve adversarial performance no worse than adversarially training the classifiers. On average, \sysshort improves a CIFAR-10 classifier's natural and adversarial accuracy by $8.20\%$ and $6.17\%$ respectively compared to adversarial training.  

Finally, we show that applying a regularization term to the classification loss allows \sysshort to achieve \textit{black-box compatibility} and thus creating a \textbf{fully model-agnostic defense} for simple datasets. Against a PGD adversary, our approach can improve the adversarial accuracy of a naturally pre-trained model for which we only have black-box access (\textit{transfer} classifier) by up to $85.33\%$ and $45.34\%$ on the MNIST and Fashion-MNIST datasets, respectively. For a classifier for which we have white-box access (\textit{native} classifier), using \sysshort slightly outperforms adversarially training the classifier by $2.32\%$ and $6.13\%$ on MNIST and Fashion-MNIST, respectively. Table~\ref{tab:properties} provides an overview of \sysshort how it compares with related work. 

\paratitle{Our Contributions}

\begin{itemize}[nosep]
    \item We propose \sys (\sysshort), a model-agnostic adversarial defense robust to a certain class of adaptive adversaries. We adversarially train an AE in order to disentangle adversarial robustness and classification so as to provide adversarial robustness for multiple classifiers.
    \item On MNIST and CIFAR-10, we show that \sysshort is able to improve the adversarial robustness of a model by $18.21\%$ and $8.76\%$ respectively compared to standard adversarial training without modifying the model.
    \item Using transferability, \sys can provide a fully model-agnostic defense improving the adversarial accuracy with respect to the PGD $\mathbf{L}_{\infty}$ attack on an MNIST and Fashion-MNIST black-box classifier by $85.33\%$ and $45.34\%$ respectively. With respect to the CW $\mathbf{L}_{2}$ attack, the adversarial accuracy is improved by 76.61\% and 62.67\% respectively.
    \item We measure the natural image corruption robustness of \sysshort on CIFAR-10 and show that it has a higher resistance to natural image corruptions than adversarially trained classifiers, which, based on previous work, indicates true adversarial robustness improvements, rather than unjustified gains due to gradient shattering.
    
    
\end{itemize}


\section{Background} 
\label{sec:bg}
\paratitle{Adversarial Attacks} Adversarial examples were introduced by Szegedy \etal~\cite{szegedy2014intriguing}. They showed that it was possible to generate \textit{imperceptible} non-random noise that, when added to the input of a deep neural network, would cause an arbitrary change in its output. Since then, numerous adversarial attacks have been developed which can be broadly classified based on the adversary's knowledge of the model: (1)~\textit{White-box attacks} assume the adversary has perfect knowledge of the parameters of the target model. Some well known white-box attacks include the Jacobian-Based Saliency Map (JSMA) attack, which expresses the forward derivatives of the target model in the form of the Jacobian matrix to generate adversarial perturbations~\cite{papernot2016limitations}, and the Fast-Gradient Sign~(FGSM) attack and its variants, which use back-propagated gradients to find an input that maximizes a given loss function~\cite{goodfellow2014explaining,kurakin2016adversarial}. (2)~\textit{Black-box attacks} assume the adversary only has the ability to query a model for its soft (probability distribution) or hard (predicted label) output. In these attacks, the adversary queries the target model and uses the responses to estimate gradient information, which can then be used to re-enact a white-box attack~\cite{Narodytska2017SimpleBA,zoo,Bhagoji2018PracticalBA,Cheng2018QueryEfficientHB}. Alternatively, it is known that adversarial inputs are transferable. An adversarial input created to cause misclassification errors on one model can be reused to cause a similar effects on other models trained on the same data despite differences in the model architectures~\cite{szegedy2014intriguing}. Using this property, an adversary can create adversarial inputs using white-box attacks on a surrogate model and then use these inputs against the black-box target model~\cite{papernot2016transferability,liu2016delving}.

\paratitle{Adversarial Training} Given the widespread use of machine learning, successful adversarial attacks against deployed systems could result in dire real-world consequences. As such, it is critical to develop techniques to achieve adversarial robustness. Madry~\etal~\cite{madry2018towards} define adversarial robustness through the following saddle-point problem:
\begin{align}
    \min_\theta\rho(\theta), \quad \text{where} \quad \rho(\theta) = \mathds{E}_{(x,y) \sim \hat{p}_{train}} \Big[ \max_{\delta \in S} \mathbf{L}(F_{\theta}(x + \delta), y)\Big]
\label{eq:saddle_point}
\end{align}
where, $\theta$ are the parameters of a neural network classifier $F$ which are updated to minimize the cross-entropy objective $\mathbf{L}$ on an input $x \in \mathds{R}^d$ and its corresponding label $y \in \{1 \cdots k\}$ drawn from the training data distribution $\hat{p}_{train}$. The budget of the adversary is defined through $S \in \mathds{R}^d$, which is the set of all possible imperceptible perturbations applicable to the input $x$.
While the inner maximization problem can be interpreted as the objective of an adversary, the outer minimization problem is a formal representation of the defender's objective. Several techniques have been proposed to find an optimal solution for the saddle-point problem defined in Eq \ref{eq:saddle_point}. Madry \etal show that certifiable robustness can be achieved by training a model against its 'universal first-order adversary' computed using the Projected Gradient Descent (PGD) attack, an iterative form of the FGSM attack. Referred to as adversarial training, Madry \etal demonstrated that this training process can create MNIST and CIFAR classifiers with significant adversarial robustness to certain adversarial inputs. Due to scalability issues on larger datasets, several improvements, such as using the single-step FGSM attack instead of PGD~\cite{kurakin2017atscale} and ensemble adversarial training~\cite{tramer2018ensembletraining}, have been proposed to reduce the performance overhead of adversarial training. 

In a later work, Ilyas~\etal developed a robust feature extraction methodology using adversarial training~\cite{Ilyas2019AdversarialEA}. The penultimate layer of an adversarially trained model is used to disentangle robust and non-robust features. Afterwards, the original dataset is projected into the robust feature space and a new classifier is naturally trained on the robust dataset. The robust feature extractor can be seen as a pre-processing defense, but we identify a few limitations compared to our proposed defense. First, it is necessary to retrain a classifier on the robust dataset in order to achieve reasonable performance. Thus, the robust feature extractor cannot be used with other classifiers unless additional training is performed. Second, the performance of the classifier on the robust dataset decreases significantly. In fact, both the natural and adversarial performance of their method is worse than the adversarially trained classifier used to build the extractor.


\paratitle{Denoising Defenses} As adversarial examples are typically generated by adding noise to a correctly classified input, it is natural to attempt to use denoising algorithms to remove the adversarial noise. Denoising the adversarial noise before classification can allow for adversarial robustness to be separate from a model's specifications. Gu and Rigazio explored using simple denoising autoencoders (AEs) as a pre-processing step to filter out the adversarial noise and re-project the input to the natural data manifold ~\cite{DBLP:journals/corr/GuR14}. Other works treat adversarial samples as a subset of out-of-distribution samples and deploy AEs to detect them based on the deviation of their reconstructions from the natural data manifold~\cite{meng2017magnet}. These works use a network architecture formed by stacking an AE with a classifier for the purposes of denoising or detecting adversarial inputs. However, such defenses assume the adversary is \textit{oblivious} to the defense. An \textit{adaptive} adversary attacking the end-to-end pipeline rather than the classifier alone is able to bypass these defenses. Liao~\etal proposed two modifications to improve the robustness of the stacked architecture defense~\cite{Liao2018DefenseAA}. First, the DAE was changed to output inverse adversarial noise to correct the modified input rather than fully denoise the original input. This change was based on the hypothesis that learning the adversarial noise added to an input is an easier task than learning to reconstruct the original input. Second, they replaced the pixel-based reconstruction loss with a loss based on the L1 distance between the intermediate layer features of a natural and an adversarial input. While these modifications to the original architecture improve the robustness of the model against an oblivious adversary, they still fail in the presence of an adaptive adversary. \sysshort, in contrast, improves the adversarial robustness against certain class of white-box adaptive adversaries in a model-agnostic manner.

Previous work has also explored using Generative Adversarial Networks (GANs) as a substitute for AEs in adversarial defenses ~\cite{defensegan,ilyas2017robust}. These defenses leverage the superior generative capacity of GANs to transform 
the adversarial image into an image that lies on the data manifold learnt by the generator. These techniques, however, have their own drawbacks. \eg it is possible to identify adversarial samples that lie on the data manifold learnt by the generator of the GAN~\cite{athalye2018obfuscated,jalal2017robust}. Also, these techniques have not been shown to be particularly effective with more complex image datasets.

\section{\sys}
\label{sec:design}

In this section, we provide a high-level overview of \sys (\sysshort). \sysshort combines model-specific adversarial training with denoising AEs to take the first step toward designing a model-agnostic defense. Recall that we define a model agnostic defense as one that when deployed for a set of models (1) is \textit{unintrusive}: It does not require modification of any model's state such as the network weights or architecture, (2) is \textit{transferable}: It can be transferred between each model in the set and achieve similar levels of adversarial robustness, and (3) is \textit{black-box compatible}: It does not rely on information about a model's state beyond its output.

To achieve the first property, \sysshort stacks a denoising AE in front of a pre-trained classifier\footnote{In all of our experiments, the pre-trained classifier attached to the AE was naturally trained with no adversarial defenses deployed} and uses adaptive adversarial training to create an end-to-end adversarially robust model with respect to first order adversaries. As the AE does not change the shape of the input, we can treat it as a preprocessing module that exists independent from the classifier. Thus, to improve the classifier's adversarial robustness,  \sysshort \textit{only} needs to modify the weights belonging to the AE, leaving the pre-trained classifier \textit{unmodified}. Let $F_{\theta}$ be the classification function learnt by the pre-trained classifier, parameterized by $\theta$. We assume white-box access to $F_{\theta}$ and so refer to it as the \textit{native classifier}. We stack a convolutional AE $G_{\phi}$, parameterized by $\phi$, in front of $F_{\theta}$ and form an augmented classification pipeline $\hat{y} = F_{\theta}(G_{\phi}(x))$. The adversarial loss $\rho(\theta)$, as defined in Eq.~\ref{eq:saddle_point}, is modified for our stacked model as follows:

\begin{equation}
    \rho(\theta,\phi) = \mathds{E}_{(x,y) \sim \hat{p}_{train}} \Big[ \max_{\delta \in S}~\mathbf{L}(F_{\theta}(G_{\phi}(x + \delta)), y)\Big]
\label{eqn:aaa_adversarial_loss}
\end{equation}

Simply put, compared to Eq.~\ref{eq:saddle_point}, the adaptive adversary now seeks to find an adversarial perturbation that maximizes cross-entropy loss of the classifier after the input has been preprocessed by the AE. Thus, we train the AE to minimize this loss.

With respect to the second property, we employ ensemble adversarial training. Ensemble adversarial training encourages the AE to learn output reconstructions that mitigate the effect of adversarial noise for each classifier in the ensemble, thus reducing the overfitting phenomenon observed in standard adversarial training by prior work~\cite{tramer2018ensembletraining}. This translates to having multiple native classifiers that the AE must make robust. Section~\ref{sec:partial_ma} provides detail on how we achieve this property and performs evaluation using the MNIST and CIFAR-10 datasets. 

In order to achieve the third property, we focus on enhancing the transferability of the AE's output. Prior adversarial attacks identified that adversarial examples are highly transferable between models. That is to say, an adversarial input designed for one model is highly likely to remain adversarial for a different model. We adopt a similar approach and train \sysshort as before, but we evaluate its robustness on a non-native classifier not seen during training (\textit{transfer} classifier). Our initial experiments demonstrated that adversarial training with traditional classification loss did not result in a transferable defense. Thus, we use reconstruction loss regularization to encourage the AE to generate output reconstructions that lie on the natural data manifold and hence, generalize to others models trained on the same input domain. In Section~\ref{sec:full_ma}, we demonstrate how this loss function allows for black-box compatibility on simple datasets (MNIST and Fashion-MNIST). However, on CIFAR-10, the transferability breaks down due to the increase data complexity.



\section{Achieving Transferability with \sysshort} \label{sec:partial_ma}

To make \sysshort transferable across multiple pre-trained native classifiers, we use ensemble adversarial training.
In their work, Tram{\`e}r \etal proposed using ensemble adversarial training to increase the diversity of adversarial examples a classifier was trained on. In their approach, they periodically select a random classifier to generate adversarial examples for, and then train the native classifier using those examples. They demonstrate that this approach reduces the overfitting problem of standard adversarial training, so we adopted their methodology. During training, we periodically randomly select a classifier as the native classifier and then update the AE to minimize the adversarial loss (Eq~\ref{eqn:aaa_adversarial_loss}) corresponding to the selected classifier. As we demonstrate later, this approach enables \sysshort to satisfy the second property of a model-agnostic defense. 

We note that while ensemble adversarial training improves diversity of adversarial examples \sysshort observes, it does not address all of the shortcomings of standard adversarial training. Namely, standard adversarial training is only robust to the class of adversarial examples used during training. Other works have identified that adversarially trained models remain vulnerable to other classes of adversarial examples such as those generated through black-box attacks~\cite{Narodytska2017SimpleBA,zoo,Bhagoji2018PracticalBA,Cheng2018QueryEfficientHB}. We leave improving the robustness of \sysshort to all known class of attacks as future work.

\paratitle{Experimental Setup}
We conduct experiments on the MNIST and CIFAR-10 datasets.
For the MNIST experiments, \sysshort uses a simple convolutional AE. We use the following three different pre-trained classifiers as a part of the ensemble: (A)~the classifier with architecture as provided by Madry~\etal~\footnote{\url{https://github.com/MadryLab/mnist_challenge}}; (B)~a simple CNN with a single convolution layer and a couple of fully connected layers; and (C)~a classifier that only contains fully connected layers. For the CIFAR-10 experiments, \sysshort uses a U-Net AE architecture~\cite{unet}. The main difference between a U-Net AE and a standard convolutional AE is the use of skip connections, which are forward feed connections between the encoding and decoding layers in the network that enable high fidelity reconstructions. Again, we use three different pre-trained classifiers as a part of the ensemble: (D)~the ResNet architecture provided by Madry~\etal;\footnote{\url{https://github.com/MadryLab/cifar10_challenge}} (E)~the VGG-19 classifier~\cite{vgg}; and (F)~a simple DNN consisting of four convolution layers and a fully connected layer. We have chosen our classifier architectures in a way that demonstrates that our defense is agnostic to the classifier's architecture. 

We use the Adam optimizer to adversarially train the AE, while keeping the weights of the native pre-trained classifiers frozen. For MNIST we generate adversarial examples at each training iteration using a 40-step L$_{\infty}$ bounded PGD attack with a step size of $0.01$ and $\epsilon=0.3$. For CIFAR-10, we generate adversarial axamples using a 10-step L$_{\infty}$ bounded PGD attack with a step size of $\frac{2}{255}$ and $\epsilon=\frac{8}{255}$.\footnote{We scale the pixel values to the range $[0,1]$} For all experiments, we set the initial learning rate at $0.001$ and decrease it if the validation loss did not improve over five epochs.

\paratitle{Evaluation} \label{subsec:partial_ma_results}
We evaluate \sysshort using two different metrics to demonstrate that the benefits of \sysshort are shared across all the classifiers in the ensemble. First, we demonstrate the adversarial robustness of \sysshort against an adaptive white-box PGD adversary (perturbation robustness). Then, we show that \sysshort improves the robustness of its native classifiers to natural image corruptions (corruption robustness), using the CIFAR-10-C dataset. Previous work has identified corruption robustness as a critical criteria for evaluating overall model robustness~\cite{pmlr-v97-gilmer19a}. 


\paratitle{Perturbation Evaluation}
For MNIST, we evaluate all models against a  $100$-step L$_{\infty}$ bounded PGD attack with $\epsilon=0.3$ and step size $a=0.01$, with 50 random restarts. For CIFAR-10, we evaluate all models against a  $20$-step L$_{\infty}$ bounded PGD attack with $\epsilon=\frac{8}{255}$ and step size $a=\frac{2}{255}$, with 10 random restarts. 

\begin{table}[h]
\vspace{-1em}
\begin{center}
\resizebox{\columnwidth}{!}{%
  \begin{tabular}{@{}cccccccc@{}}
    \toprule
    \multicolumn{4}{c}{\textbf{MNIST}} & \multicolumn{4}{c}{\textbf{CIFAR-10}}\\\cmidrule(r){1-4} \cmidrule(r){5-8}
    \textbf{Model} & \textbf{Defense} & \textbf{Natural} & \textbf{PGD} & \textbf{Model} & \textbf{Defense} & \textbf{Natural} & \textbf{PGD}\\
    \cmidrule(r){1-4} \cmidrule(r){5-8}
    \multirow{2}{*}{A} & AT & 99.10\% & 88.81\% & \multirow{2}{*}{D} & AT & \textbf{81.49\%} & \textbf{45.89\%} \\
     & AAA  & 99.10\% & \textbf{90.29\%} & & AAA  & 80.06\% & 45.34\% \\
    \cmidrule(lr){1-4} \cmidrule(lr){5-8}
    \multirow{2}{*}{B} & AT & 98.37\% & 87.67\% & \multirow{2}{*}{E} & AT & 71.95\% & 39.73\% \\
     & AAA  & \textbf{99.19\%} & \textbf{90.12\%} & & AAA & \textbf{78.22}\% & \textbf{42.94\%} \\
    \cmidrule(lr){1-4} \cmidrule(lr){5-8}
    \multirow{2}{*}{C} & AT & 93.32\% & 71.76\% & \multirow{2}{*}{F} & AT & 69.48\% & 36.40\% \\
     & AAA  & \textbf{99.17\%} & \textbf{89.97\%} & & AAA & \textbf{79.60}\% & \textbf{45.16\%} \\
    \bottomrule
  \end{tabular}
}
\caption{Perturbation evaluation for MNIST and CIFAR-10 datasets. \sysshort successfully protects all classifiers in the ensemble and has comparative or better performance compared to traditional adversarial training.}
\label{tab:ensemble_results}
\end{center}
\vspace{-1.5cm}
\end{table}

Table \ref{tab:ensemble_results} compares the robustness benefits of using \sysshort with a pre-trained classifier against re-training the classifier using traditional adversarial training (AT). Our results show that \sysshort has comparable or better performance than traditional adversarial training when evaluated against the $\mathbf{L}_{\infty}$ PGD attack. For models B, C, E, and F (weaker classifiers in the respective ensembles), \sysshort significantly improved both the natural and adversarial accuracy  (\eg an additional $10.12\%$ and $8.76\%$ for Model F's natural and adversarial accuracy on CIFAR-10, respectively). Thus for a set of pre-trained classifiers, \sysshort protects all the classifiers in the ensemble. With traditional adversarial training, each classifier would need to be individually re-trained adversarially.


\paratitle{Corruption Evaluation}
Recently, Gilmer~\etal formally recognized a relationship between adversarial robustness and corruption robustness~\cite{pmlr-v97-gilmer19a}. They showed that adversarial examples are a natural phenomenon due to the fact that trained models have non-zero test error on natural image corruptions. Thus, adversarial robustness can be improved by improving a model's robustness to natural image corruptions and vice versa. Using the corruption dataset provided by Hendrycks and Dietterich~\cite{corrpution}, they show that adversarial training, a proven defense method, significantly improves a model's robustness to Gaussian noise corruptions as well as other forms of corruption. Furthermore, they show that other defense methods, which employ vanishing gradient strategies, do not result in any significant improvement on Gaussian noise corruptions. Based on their results, they suggest that future defense strategies should evaluate performance on corrupted image distributions as a sanity check to ensure that vanishing gradients is not the main explanation for improved adversarial robustness.

\begin{figure*}[h]
\vspace{-2em}
\includegraphics[width=\columnwidth]{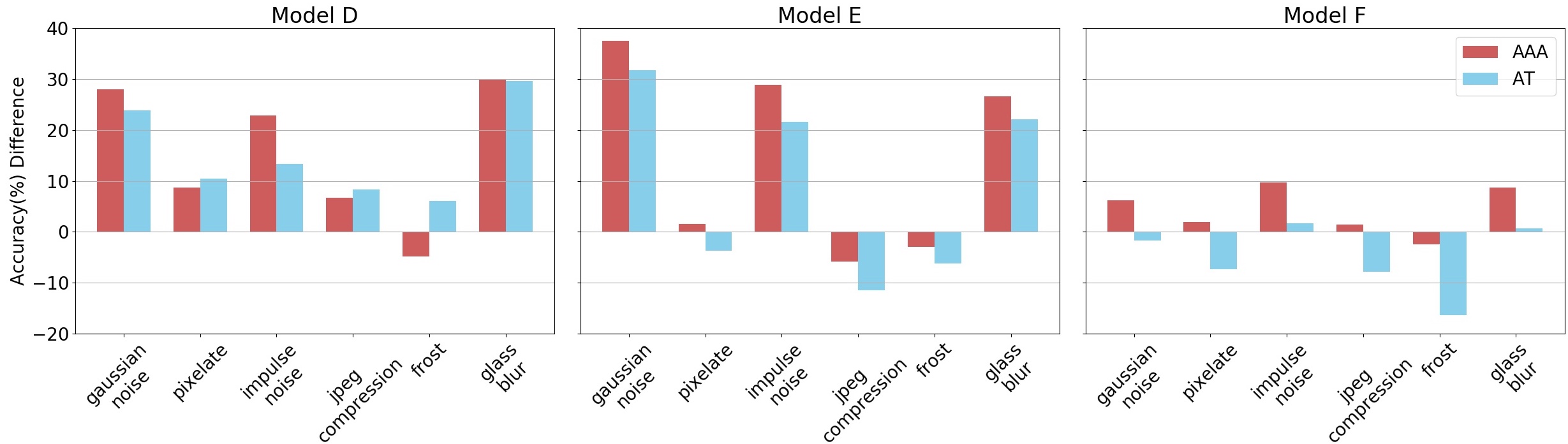} 
\caption{Comparing corruption robustness of Adversarial Trained classifiers (AT) to pre-trained classifiers protected by \sysshort. The \sysshort in this case has been trained using the ensemble of all three pre-trained classifiers. The $y$-axis represent difference in accuracy w.r.t. pre-trained version of the classifier. In case of the strongest classifier in our ensemble (ResNet), we observe comparable performance between AT and \sysshort. However, for the weaker classifiers (VGG and DNN), we observe increased robustness for \sysshort as compared to AT.}
\label{fig:cifar_results_4}
\vspace{-0.6cm}
\end{figure*}

Inspired by this, we use the the corruption test set for CIFAR-10 referred to as CIFAR-10-C~\cite{corrpution} and evaluate the corruption robustness of \sysshort. We compare the performance of \sysshort and adversarial trained classifiers and present these results in Figure \ref{fig:cifar_results_4}. Note that only results for 5 of the 19 possible corruptions are shown, but additional figures and the complete results can be found in the Supplementary materials.

The \sysshort protected pre-trained Model D has comparable performance to an adversarially re-trained Model D. Also, the \sysshort protected pre-trained Models E and F exhibit much greater robustness to natural image corruptions compared to their adversarially re-trained versions. Furthermore, in cases where adversarial training decreases the corruption robustness, \sysshort either has a lesser decrease or a positive increase relative to the baseline natural classifier. Based on these results and prior work, \sysshort appears to not rely on gradient shattering to improve adversarial robustness.

\section{Achieving Black-box Compatibility with \sysshort} \label{sec:full_ma}
In this section, we enhance the transferability of \sysshort to satisfy the third property of a model-agnostic defense. While using the cross-entropy loss was sufficient to achieve transferability across the ensemble of native classifiers, it caused the AE to overfit to these classifiers and only improve their individual adversarial robustness. This led to poor initial results when transferred to a non-native (transfer) classifier. Upon further examination, we observed that the reconstructions of the AE trained using the cross-entropy loss were visually unrecognizable (Table \ref{tab:recon}). We speculate that these reconstructions are features lying on an abstract feature space, that result in good performance specific to the native classifiers.

\begin{table*}[h!]
\vspace{-2em}
    \centering
    \resizebox{\textwidth}{!}{%
    \begin{tabular}{@{}M{25mm}M{25mm}M{25mm}M{25mm}@{}}
    \toprule
    Input & \lossx & \lossm & \lossxm   \\
   \midrule
    \includegraphics[width=0.15\textwidth]{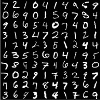} &     \includegraphics[width=0.15\textwidth]{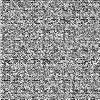} &     \includegraphics[width=0.15\textwidth]{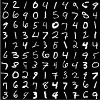} &     \includegraphics[width=0.15\textwidth]{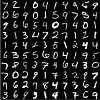}\\
    
    \includegraphics[width=0.15\textwidth]{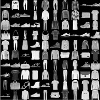} &     \includegraphics[width=0.15\textwidth]{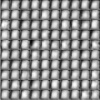} &     \includegraphics[width=0.15\textwidth]{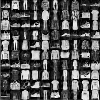} &     \includegraphics[width=0.15\textwidth]{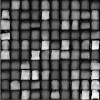}\\
    
    \includegraphics[width=0.15\textwidth]{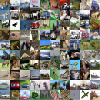} &     \includegraphics[width=0.15\textwidth]{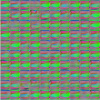} &     \includegraphics[width=0.15\textwidth]{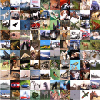} &     \includegraphics[width=0.15\textwidth]{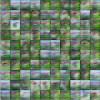}\\

    \bottomrule
    \end{tabular}
    }
\caption{Visualization of the output of the AE trained using cross-entropy loss (\lossx), reconstruction loss (\lossm), and their combination (\lossxm). The addition of \lossm to our training objective visually balances the abstract representations learned by \lossx.}
\label{tab:recon}
\vspace{-1cm}
\end{table*}

Thus, to regularize the AE so it does not overfit to the native classifiers, we added a standard reconstruction loss as an additional loss term.  In Table \ref{tab:recon}, we see that the additional term serves to balance the abstract representation learned by training on cross-entropy only, pushing them closer to the natural data manifold.

Formally, let $\mathds{\widetilde{B}}$ be a batch of $m$ adversarial images $\{\mathbf{\widetilde{x}}^{(1)}, \cdots, \mathbf{\widetilde{x}}^{(m)}\}$ corresponding to batch $\mathds{B} = \{\mathbf{x}^{(1)}, \cdots, \mathbf{x}^{(m)}\}$ of natural images sampled from $\hat{p}_{train}$. We obtain $\mathds{\widetilde{B}}$ from $\mathds{B}$ by solving the maximization problem in Eq.~\ref{eqn:aaa_adversarial_loss} using PGD. Let $J(\mathds{\widetilde{B}}; \mathbf{\theta}, \mathbf{\phi})$ be a cost function for the mean of the adversarial loss defined in Eq.~\ref{eqn:aaa_adversarial_loss} computed over the batch $\mathds{\widetilde{B}}$. Our training objective updates AE parameters $\mathbf{\phi}$, while keeping classifier parameters $\mathbf{\theta}$ constant, so as to minimize the following:
\begin{equation}
    J(\mathds{\widetilde{B}}; \mathbf{\theta}, \mathbf{\phi}) + \lambda \dfrac{1}{m}\sum_{i=1}^{m} \Big( G_{\mathbf{\phi}}(\mathbf{\widetilde{x}}^{(i)}) - \mathbf{x}^{(i)} \Big)^2
\label{eqn:batch_loss}
\end{equation}

where, $\lambda$ is a constant that we use to balance the magnitudes of the two losses. In the remainder of the paper, we refer adversarial cross-entropy loss, reconstruction loss, and their combination (Eq. \ref{eqn:batch_loss}) as \lossm, \lossx, and \lossxm respectively.



\paratitle{Experimental Setup}
To demonstrate that \sysshort is black-box compatible, we replace the native classifier in our pipeline, at test time, with another classifier which was independently trained on the same data as the native classifier. We refer to this classifier as the \textit{transfer} classifier and we chose it to have considerably different architecture than its native counterpart. Note that we only have black-box access to the transfer classifier. To evaluate the robustness of our defense, we attack both the native and the transfer \sysshort models in an end-to-end manner using the PGD attack and demonstrate performance of \sysshort model on the MNIST and Fashion-MNIST datasets~\cite{mnist,fashion-mnist}. We do not include CIFAR-10 since our experiments didn't yield a robust model-agnostic defense, indicating further improvements are required.

For the MNIST experiments, we remove Model C from the previous section from the native ensemble and treat it as the transfer classifier. For the Fashion-MNIST experiments, we use the classifier provided by Zheng~\etal~\footnote{\url{https://github.com/tianzheng4/Distributionally-Adversarial-Attack}} as the native classifier and a classifier containing only fully-connected layers as the transfer classifier.

We use the Adam optimizer to adversarially train the AE using the three loss functions, \ie \lossx, \lossm, and \lossxm, while keeping the weights of the native classifier frozen. We generate adversarial examples at each training iteration using a 40-step L$_{\infty}$ bounded PGD attack with a step size of $0.01$ and $\epsilon=0.3$ and $\epsilon=0.2$ respectively. We set the initial learning rate at $0.001$ and decrease it if the validation loss does not improve over five epochs.

\paratitle{Evaluation}
We evaluate all models against two white-box attacks. For the first attack, we used a $100$-step  L$_{\infty}$ bounded PGD attack with step size of $0.01$ and $\epsilon=0.3$ and $\epsilon=0.2$ respectively. We perform 50 random restarts for each input so an input is only considered correctly classified if all 50 adversarially modified versions are correctly classified. For the second attack, we used the Carlini-Wagner (CW) attack \cite{Carlini2016TowardsET}. The attack finds an adversarial perturbation $\delta$ such that:

\begin{equation}
    \min_{\delta} \quad \mathcal{D}(x,x+\delta) + c * \mathbf{L}(F_{\theta} (x+\delta))
\end{equation}

where $\mathcal{D}$ is the distance function and $\mathbf{L}$ is the loss function representing classification performance. In our experiments, we used $\mathbf{L}_2$ version of the attack with c=100 targeting logit loss based on previous works \cite{defensegan}. We compare the performance of adversarial training to our approach in Tables~\ref{tab:mnist_results} and~\ref{tab:fmnist_results}. 

\begin{table*}[h]
\begin{center}
\resizebox{\columnwidth}{!}{%
\begin{tabular}{@{}ccccccc@{}}
\toprule
\multirow{2}{*}{\textbf{Models}} & \multicolumn{3}{c}{\textbf{Native Classifier}} & \multicolumn{3}{c}{\textbf{Transfer Classifier}}\\
\cmidrule(lr){2-4} \cmidrule(lr){5-7}
& \textbf{Natural} & \textbf{PGD} & \textbf{CW-$\mathbf{L}_2$} & \textbf{Natural} & \textbf{PGD} & \textbf{CW-$\mathbf{L}_2$} \\
\midrule
Natural Training & 99.24\% & 0\% & 0\% & 98.00\% & 0\% & 0\% \\
Adversarial Training & 99.10\% & 88.81\% & 52.10\% & 93.32\% & 71.76\% & 26.64\% \\
\midrule
\sysx & 99.08\% & 90.44\% & 58.16\% & 28.75\% & 11.85\% & 1.94\% \\
\sysm & 99.17\% & 89.20\% & 65.86\% & 98.21\% & 82.17\% & 57.16\% \\
\sysxm & 99.15\% & \textbf{91.13\%} & \textbf{74.44\%} & 98.12\% & \textbf{85.33\%} & \textbf{76.61\%} \\
\bottomrule
\end{tabular}
}
\end{center}
\caption{Comparing adversarial robustness of \sysshort to the adversarial training approach proposed by Madry~\etal~\cite{madry2018towards} on MNIST dataset. The PGD attack is a 100-step attack with 50 random restarts. The CW-$\mathbf{L}_2$ attack uses c=100 and adversarial examples are successful if $\mathbf{L}_2$ distance is less than 3. Transfer classifiers were chosen such as to offer variability compared to the architecture of the native classifier. \sysshort appears to be fully model-agnostic.}
\label{tab:mnist_results}
\vspace{-2em}
\end{table*}

\begin{table*}[h]
\begin{center}
\resizebox{\columnwidth}{!}{%
\begin{tabular}{@{}ccccccc@{}}
\toprule
\multirow{2}{*}{\textbf{Models}} & \multicolumn{3}{c}{\textbf{Native Classifier}} & \multicolumn{3}{c}{\textbf{Transfer Classifier}}\\
\cmidrule(lr){2-4} \cmidrule(lr){5-7}
& \textbf{Natural} & \textbf{PGD} & \textbf{CW-$\mathbf{L}_2$} & \textbf{Natural} & \textbf{PGD} & \textbf{CW-$\mathbf{L}_2$} \\
\midrule
Natural Training & 92.55\% & 5.40\% & 0\% & 89.74\% & 4.33\% & 0\%\\
Adversarial Training & 87.73\% & 62.16\% & 49.43\% & 79.11\% & \textbf{61.28\%} & 24.50\% \\
\midrule
\sysx & 87.59\% &\textbf{68.29\%} & 65.36\% & 33.60\% & 22.54\% & 31.32\%  \\
\sysm & 89.26\% & 21.87\% & 55.67\% & 88.38\% & 28.71\% & 54.83\% \\
\sysxm & 87.40\% & 63.98\% & \textbf{69.35\%} & 74.44\% & 49.67\% & \textbf{62.67\%} \\
\bottomrule
\end{tabular}
}
\end{center}
\caption{Comparing adversarial robustness of \sysshort to the adversarial training approach proposed by Madry~\etal~\cite{madry2018towards} on Fashion-MNIST dataset. The PGD attack is a 100-step attack with 50 random restarts. The CW-$\mathbf{L}_2$ attack uses c=100 and adversarial examples are successful if $\mathbf{L}_2$ distance is less than 3. Transfer classifiers were chosen such as to offer variability compared to the architecture of the native classifier. \sysshort appears to be fully model-agnostic.}
\label{tab:fmnist_results}
\vspace{-3em}
\end{table*}

For all three loss functions, \sysshort minimally impacts the natural accuracy of the pre-trained native classifier, while significantly improving the adversarial accuracy with respect to both PGD and CW adversarial examples. \sysm has the worst PGD adversarial robustness, especially on Fashion-MNIST, likely due to the fact that the AE is trained without taking into account the classification loss.\footnote{We attribute the high adversarial accuracy of \sysm on MNIST to the simple nature of the dataset. In many studies, it has been shown that the simplicity of MNIST allows for defense solutions that do not scale to larger datasets} We also see that \lossm helps to improve the model's robustness to CW adversarial inputs despite poor adversarial robustness against PGD attacks.


Unlike our native results, we see that \sysx has extremely poor adversarial robustness. This drop in performance on transfer is likely due to the overfitting of the reconstructed output to the native classifier. In Table~\ref{tab:recon}, we see that the reconstructions for \lossx are not recognizable to a human, which suggests that they may not generalize well across pre-trained classifiers as well. As we expected, \lossm encourages better transferability performance than \lossx as it is an unsupervised loss objective that serves to regularize the training objective and obtain more generalizable outputs. We see in Table~\ref{tab:recon}, the reconstructions created when using \sysm appear very similar to natural input samples. However, as it has poor native adversarial robustness against PGD attacks, the transferability benefits are limited. By combining the two loss functions, we see that \sysxm has the best adversarial performance when transferred to a new classifier. \lossx serves to improve the robustness to a PGD attack and \lossm serves to both improve the robustness to a CW attack and regularize the reconstructions to enable transferability. The improvement in transferability performance can be attributed to the idea that classifiers learn similar decision boundaries with respect to the natural data~\cite{goodfellow2014explaining}. Note that \textbf{no additional training on \sysshort was performed} to protect the transfer classifier.
Unfortunately, this approach is not sufficient to provide a fully transferable defense on CIFAR-10. In this case, we observed that even though exhibiting comparable robustness on the native classifier, \sysshort was unable to transfer robustness benefits to the transfer classifiers. In Section~\ref{sec:discussion}, we discuss future directions that may allow a fully model-agnostic defense for complex datasets.


\section{Comparison with other generative model based defenses}
In this section, we provide a detailed comparison between \sysshort and prominent prior defense methods that make use of generative models (\eg AEs and GANs). We refer to prior works that break these defense methods, and demonstrate the robustness of \sysshort against these attacks where applicable. The main advantage of \sysshort  compared to the aforementioned methods is that \sysshort is robust against adaptive white-box PGD attacks.

\paratitle{High-Level Representation Guided Denoiser~\cite{Liao2018DefenseAA}}
Liao~\etal make use of a denoising AE that borrows its design from the U-Net architecture to attain the unintrusiveness property of a model-agnostic defense. They empirical validate that their defense method satisfies the other two properties. The AE, however, is trained against a static set of adversarial examples that is pre-generated against an ensemble of naturally trained classifier. Athalye \etal~\cite{athalye2018robustness} show that this defense can be broken (with $100\%$ success rate) by applying the PGD attack to the end-to-end AE-classifier pipeline. \sysshort on the other hand has been trained to be robust against such an adversary and therefore doesn't break under such an attack. 


\paratitle{MagNet~\cite{meng2017magnet} and DefenseGAN~\cite{defensegan}}
Both these defenses train a generative models (AE and GAN, respectively) to pre-process the input independent of the classifier and re-project the input back on the natural data manifold. This allows them to attain the unintrusiveness property. Also, they train their defense in a way that doesn't take into account the classifier or any specific set of adversarial examples, which enables their defense to attain the second and third properties. Both these defenses have however been broken using adaptive attacks. MagNet claims robustness in a gray-box setting where the adversary is aware of everything about the AE except its weights. Even in this setting, Carlini and Wagner \cite{carlini2017magnet} demonstrate that this defense can be broken (with $99\%$ success rate on MNIST and $100\%$ on CIFAR-10) using a transferable CW $\mathbf{L}_2$ attack. As part of our evaluations we also directly attack \sysshort with the CW $\mathbf{L}_2$ attack, and demonstrate robustness (Table~\ref{tab:mnist_results} and \ref{tab:fmnist_results}). For DefenseGAN, Jalal \etal \cite{jalal2017robust} design a latent space attack, which they call the 'overpowered attack', that breaks it with a success rate of $97\%$. In this case, the attack used was tailored to work on this specific defense, as opposed to more general adaptive attacks that worked on the other two defenses discussed in this section. We perform two different types of attacks on \sysshort that focus on the AE part: (1)~Using \lossm as a part of the PGD adversary's loss; (2)~Updating the input to the AE such that the corresponding latent space vector is as close to the latent space vector of a different class input as possible. Both these attacks are ineffective in breaking \sysshort. These results can be found in the Supplemental material.

\section{Future Improvements}
\label{sec:discussion}


 As noted in Section \ref{sec:design}, \sysshort is not a fully model-agnostic defense because it relies on the model's loss gradients in order to generate adversarial examples and update the weights of the AE. However, relying on this information limits the use of existing proposed defenses, including ours. As described earlier, the user and designer of a machine learning model may not be the same entity, which means that modifications to the model's state  may not be possible due to a lack of visibility into model (\eg third party model reuse), sensitive model information (\eg proprietary resources or sensitive training data), or insufficient resources to train a new robust model from scratch. In such scenarios, it is key to be able to deploy a model-agnostic defense, which treats the protected model as a black box, both during training and testing.
 
 To this end, we used transferability to achieve a fully model-agnostic defense. Our results in Section \ref{sec:full_ma} were promising, demonstrating that on simple datasets like MNIST and Fashion-MNIST, reconstruction loss regularization and ensemble adversarial training were sufficient to create a transferable AE defense. However, these results did not apply when measured on CIFAR10.
 
 Transferabililty, however, is only the one technique adversarial attacks use when targeting models they have black-box access to. An alternative approach used in these cases is gradient estimation. With respect to adversarial attacks, the adversary is assumed to have access to a query interface, which returns the classifier's predicted label or probability distribution for a given input. Prior black-box attacks have used estimation techniques such as the finite difference method to estimate a loss gradient and a perform normal white-box adversarial attack~\cite{zoo,Bhagoji2018PracticalBA}. This approach appears promising with respect to adapting \sysshort into a fully model-agnostic defense. Rather than use the exact loss gradient of the model, we would estimate the model gradient using the finite difference method. First, we use the existing black box adversarial attacks to generate adversarial training examples. Then, we use the finite difference method to approximate the classifier's loss gradient with respect to the output of the AE. Finally, as we have white-box access to the AE, we chain the estimated gradient through the AE and update the AE parameters accordingly.

\section{Conclusion}
In this paper, we propose \sys as a first step toward developing model-agnostic adversarial defenses. \sysshort provides gains in robustness comparable to, and in some cases better than, traditional adversarial training across a variety of perturbations/corruptions. On MNIST and CIFAR-10, we used \sysshort with ensemble adversarial training to achieve a partial model-agnostic defense; creating a single AE that was transferable across an ensemble of pre-trained classifier in a white-box setting. \sysshort was able to improve the adversarial robustness of each \textbf{naturally-trained} classifier in the ensemble achieving, at a minimum, comparable adversarial performance to adversarial training. On MNIST and Fashion-MNIST, \sysshort improved the robustness against adaptive PGD $\mathbf{L}_{\infty}$ bound adversary and CW $\mathbf{L}_2$ bound adversary on a \textbf{naturally-trained} classifier. Furthermore, \sysshort was black-box compatible, able to be \textbf{transferred} to a new, never before seen classifier without any additional training to improve its adversarial robustness. Finally, based on previous work, which uncovered the relationship between a model's robustness to natural image corruptions and its adversarial robustness, we measured the natural image corruption robustness of \sysshort. Our evaluation revealed that \sysshort outperforms traditional adversarial training across multiple image corruptions.  To our knowledge, \sys represents the first model-agnostic adversarial defense that is robust to a class of adaptive adversaries and natural image corruptions.

\clearpage
%
%
\bibliographystyle{splncs04}
\bibliography{egbib}

\clearpage

\end{document}